\documentclass{article}
\usepackage{spconf,amsmath,graphicx}
\usepackage{hyperref}

\title{Instrument Classification of Solo Sheet Music Images}

\name{Kevin Ji$^*$\qquad Daniel Yang$^*$ \qquad TJ Tsai\thanks{$*$ Equal contribution.}\thanks{\copyright~2021 IEEE. Personal use of this material is permitted. Permission from IEEE must be obtained for all other uses, in any current or future media, including reprinting/republishing this material for advertising or promotional purposes, creating new collective works, for resale or redistribution to servers or lists, or reuse of any copyrighted component of this work in other works.}}
\address{Harvey Mudd College, Claremont, CA USA}

\begin{document}
\ninept
\maketitle
\begin{abstract}
This paper studies instrument classification of solo sheet music.  Whereas previous work has focused on instrument recognition in audio data, we instead approach the instrument classification problem using raw sheet music images.  Our approach first converts the sheet music image into a sequence of musical “words” based on the bootleg score representation, and then treats the problem as a text classification task.  We show that it is possible to significantly improve classifier performance by training a language model on unlabeled data, initializing a classifier with the pretrained language model weights, and then finetuning the classifier on labeled data.  In this work, we train AWD-LSTM, GPT-2, and RoBERTa models on solo sheet music images from IMSLP for eight different instruments.  We find that GPT-2 and RoBERTa slightly outperform AWD-LSTM, and that pretraining increases classification accuracy for RoBERTa from 34.5\% to 42.9\%.  Furthermore, we propose two data augmentation methods that increase classification accuracy for RoBERTa by an additional 15\%.
\end{abstract}
\begin{keywords}
instrument, classification, identification, recognition, sheet music
\end{keywords}
\section{Introduction}

It is relatively easy for a human to identify an instrument based on its sound, since most have a very distinctive timbre. This paper asks a more challenging question: “Can you distinguish between different instruments based on their \textit{sheet music}?”  Such a system could be useful for semantic segmentation of a sheet music document (e.g.~PDF from IMSLP) that contains parts from multiple instruments either on a single page (e.g.~orchestral score) or on different pages (e.g.~violin solo part followed by the piano accompaniment part).  We study this problem for 8 primarily monophonic instruments: violin, flute, clarinet, oboe, trumpet, cello, viola, and guitar. This is the instrument classification task based on raw sheet music images.

Many previous works have studied instrument identification in audio recordings.  Much of the recent work has been spurred on by OpenMIC-2018 \cite{humphrey2018openmic}, an open dataset for multiple instrument recognition.  Nearly all recent approaches use convolutional or recurrent neural networks of various forms to process the audio spectrogram \cite{han2017deep}\cite{gururani2018instrument}, often combined with other preprocessing steps like source separation \cite{gomez2018jazz}.  Many recent works also utilize transfer learning, using pretrained CNN models as a feature extractor to train the instrument classification model\cite{humphrey2018openmic}\cite{gururani2019attention}\cite{anhari2020learning}.  A very recent work demonstrates the effectiveness of a ResNet architecture on the task \cite{koutini2020receptive}.

We approach the problem of instrument classification from a different angle, focusing on sheet music images rather than acoustic sound.  Clearly, each instrument has certain characteristics in its music that go beyond acoustic timbre: the instrument's range, monophony vs polyphony, characteristic motions (e.g.~large jumps are easier to play on a violin than a trumpet), etc.  This paper explores the extent to which we can differentiate between 8 different instruments based on note characteristics in their sheet music.

Our approach to the problem is to convert sheet music into “words” and then to approach the problem as a text classification task. A similar approach was recently proposed to classify piano music by composer \cite{tsai2020composer}.  In this previous work, much of the information is contained in the vertical component since piano is a highly polyphonic instrument.  In this work we explore a similar approach to the instrument classification task, but we focus on primarily monophonic instruments where relatively little information is contained in the vertical component.

Our work takes inspiration from recent developments in the NLP community.  In 2018, Howard and Ruder \cite{howard2018universal} achieved state-of-the-art results on several NLP tasks by training an LSTM-based language model on a large set of unlabeled data, replacing the last output layer, and finetuning on a small set of labeled data.  While the idea of pretraining a model using language modeling was not new (e.g.~\cite{dai2015semisupervised}), the impressive results achieved by this study brought language model pretraining into the mainstream.  Many other approaches like BERT \cite{devlin2019bert} and GPT-2 \cite{radford2019language} adopted a similar pretraining stage with transformer-based models, and all of their extensions (e.g.~\cite{brown2020language}\cite{dai2019transformer}\cite{guu2020realm}\cite{yang2019xlnet}) similarly use pretraining on unlabeled data.

We incorporate such pretraining methods into our approach.  We start by converting each sheet music image into a sequence of words based on the bootleg score representation \cite{yang2019midi}.  We then feed this sequence of 
words into a classifier.  We show that the performance of the classifier can be improved significantly by pretraining a language model, initializing the classifier with the pretrained language model weights, and then fine-tuning the classifier on labeled data.  We train AWD-LSTM \cite{merity2018regularizing}, RoBERTa \cite{liu2019roberta}, and GPT-2 \cite{radford2019language} language models on solo instrumental sheet music obtained from the International Music Score Library Project (IMSLP) website.  Through pretraining, we are able to improve the accuracy of a RoBERTa model from \(34.5\%\) to \(42.9\%\) on our instrument classification task.

This paper has three main contributions.  First, we define the task of instrument classification based on sheet music images, and release a benchmark based on IMSLP data for evaluating performance on an 8-way classification task.  Second, we demonstrate the benefits of language model pretraining with several model architectures combined with a recently proposed feature representation of sheet music called a bootleg score.  Finally, we introduce a simple scheme of training and test time augmentation on the bootleg score feature representation that significantly improves classification performance.\footnote{Code can be found at \url{https://github.com/HMC-MIR/InstrumentID}.} 

\section{System Description}

There are three main components in our proposed method: language model pretraining, classifier finetuning, and inference.  These three main components will be explained in depth in the next three subsections.

\subsection{Language Model Pretraining}

The first main component is language model pretraining, which consists 
of three steps.

\begin{figure}
    \centerline{\includegraphics[width=0.65\columnwidth]{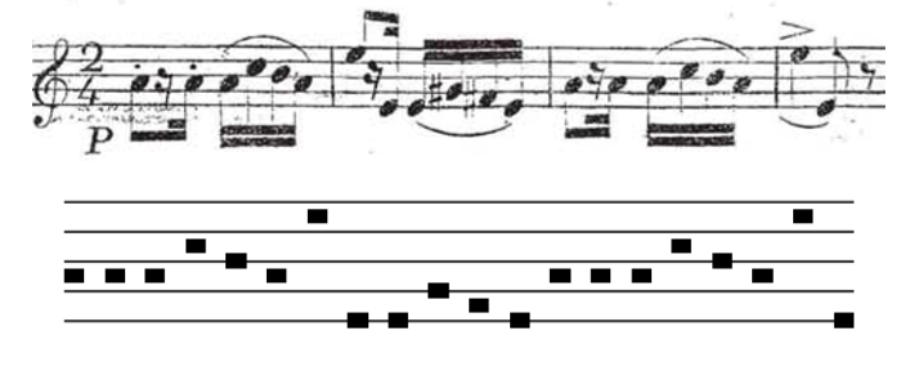}}
  
  \caption{A section of Paganini's Caprice 24 for violin and its corresponding bootleg score. The staff lines are shown for reference but are not present in the actual representation.}
  \label{fig:res1}
\end{figure}

The first step is to convert each sheet music image into a bootleg score. The bootleg score is a low-dimensional representation of sheet music that encodes the positions of filled noteheads relative to the staff lines \cite{yang2019midi}.  While the original formulation of the bootleg score was designed specifically for piano sheet music, we can easily adapt the representation to single-staff sheet music by removing the grouping of staves and adjusting the range of possible staff line positions to the eight instruments of interest.  Our modified bootleg score for single-staff sheet music is a \(31 \times N\) binary matrix, where 31 is the number of possible staff line positions and \(N\) is the number of note events in the sheet music.  Figure 1 shows an excerpt of sheet music and its corresponding bootleg score representation.

The second step is to tokenize the bootleg score into a sequence of words or subwords. We use different tokenization methods for different types of language models.  For word-based language models like AWD-LSTM \cite{merity2018regularizing}, we interpret each column of the bootleg score as a single word.  We keep any words that occur more than 3 times in the dataset, and map infrequent words to a special unknown word token \textless unk\textgreater, resulting in a vocabulary size of 4616.  For subword-based language models like RoBERTa \cite{liu2019roberta} and GPT-2 \cite{radford2019language}, the tokenization consists of two sub-steps. The first sub-step is to convert each column of the bootleg score (containing 31 bits of information) into a sequence of 4 bytes.  The second sub-step is to treat each byte as a character and learn a vocabulary of subwords in an unsupervised manner using a byte pair encoding (BPE) algorithm \cite{gage1994new}.  For the two subword-based language models in this work, we use the same BPE tokenizer with a maximum vocabulary size of 30,000.  After this step, our sheet music images have been converted into a sequence of words or subwords.

The third step is to train a language model on a set of unlabeled data. We use three different models: AWD-LSTM \cite{merity2018regularizing}, RoBERTa \cite{liu2019roberta}, and GPT-2 \cite{radford2019language}. The AWD-LSTM is a three-layer LSTM with multiple types of dropout and regularization.  The AWD-LSTM model predicts the next word in a sequence, so it can be trained in a self-supervised manner.  The second model is GPT-2, which consists of six transformer decoder layers \cite{vaswani2017attention} and is trained to predict the next token at each time step.   The third model is RoBERTa, which is an optimized version of Google's BERT \cite{devlin2019bert}.  The RoBERTa model consists of six transformer encoder layers and is trained to predict the identity of randomly masked input tokens.

\subsection{Classifier Finetuning}

The second main component is classifier finetuning, which consists of three steps.

The first step is to convert the sheet music images in the labeled dataset into sequences of words or subwords.  We follow the same process for extracting bootleg scores and use the same tokenizer that was used during language model pretraining.

The second step is to sample fixed-length fragments of words or subwords from the labeled data.  This allows us to significantly augment the amount of training data, and it also allows us to sample the data to ensure balanced classes.  By preparing the data in this manner, our classifiers are trained to classify fixed-length fragments rather 
than full pages of sheet music.  We will refer to the fragment classification problem as the proxy classification task.

The third step is to finetune a classifier on the labeled fragment data.  Our general approach is to add a classifier head on top of the pretrained language model, initialize the weights of the classifier with the pretrained language model weights, and then finetune the classifier on the labeled fragments in the proxy dataset.  For the AWD-LSTM model, the classifier head consists of two linear layers at the output of the last LSTM layer.  For the GPT-2 and RoBERTa models, we add the symbols $<$s$>$ and $<$/s$>$ at the beginning and end of every input, and we feed the output from the last transformer layer at the first (RoBERTa) and last (GPT-2) time step into a single linear classification layer.

We finetuned all models using the techniques proposed in \cite{howard2018universal}, which were shown to be effective in pretraining and finetuning text-based classifiers.  This includes using a learning rate rangefinder, gradual unfreezing of layers, discriminative finetuning, and (multiple cycles of) one cycle training \cite{smith2018disciplined}.  These methods were also found to be effective in \cite{tsai2020composer} for a composer classification task.

\subsection{Inference}
The third main component is to use the proxy classifier to make predictions on pages of unseen sheet music. We explore two different inference methods.  The first method is to convert the page into a sequence of words or subwords, and then apply the proxy classifier to a single variable-length sequence of tokens.  The second method converts the variable-length sequence of words/subwords into a sequence of fixed-length fragments with \(50\%\) overlap.  Each fragment is then processed by the proxy classifier, and the predictions are averaged across all fragments in the page.

\section{Experimental Setup}

We will explain the experimental setup in three parts: data collection, data annotation, and data preparation for the proxy task.

\begin{table*}
  \centering
  \begin{tabular}{|l|c|c|c|c|c|c|c|c|}
  \hline
  \textbf{Dataset} & \textbf{Cello} & \textbf{Clarinet} & \textbf{Flute} & \textbf{Guitar} & \textbf{Oboe} & \textbf{Trumpet} & \textbf{Viola} & \textbf{Violin} \\ \hline
  Unlabeled - Pieces    & 227   & 77       & 192   & 1513   & 82   & 85      & 84    & 520    \\ \hline
  Unlabeled - Pages     & 2264  & 1384     & 2038  & 7213   & 963  & 753     & 1408  & 5486   \\ \hline
  Labeled - Pieces      & 75    & 75       & 73    & 75     & 65   & 51      & 72    & 70     \\ \hline
  Labeled - Valid Pages & 921   & 758      & 1300  & 569    & 622  & 396     & 1323  & 1815   \\ \hline
  \end{tabular}
  \caption{Piece and page counts for each instrument in the labeled and unlabeled datasets.}
\end{table*}

The data consists of raw sheet music images from IMSLP.  The IMSLP metadata specifies the instrumentation of each piece through category tags.  Examples of these category tags include ``for violin" and ``for clarinet, piano."  We selected eight primarily monophonic instruments with a significant amount of data (shown in Table 1).  To find solo sheet music for these instruments, we filtered all IMSLP pieces based on three criteria: (1) it contained the ``for $<$instrument$>$" category tag for one of the eight instruments, (2) it was in the Public Domain or had a Creative Commons license, and (3) it was not a manuscript.  To ensure that we had enough data for each instrument, we further augmented the trumpet data with the ``for trumpet, piano" and ``for trumpet, orchestra" tags, and we augmented the oboe data with the ``for oboe, orchestra" tag.  We refer to the resulting set of filtered data as the unlabeled dataset, since the PDFs may contain accompaniment parts or filler pages.  Since there can be multiple PDFs for a given piece, we selected the PDF with the most downloads.

Next, we constructed a labeled dataset by manually annotating a subset of the unlabeled data. We first randomly sampled 75 PDFs from each instrument in the unlabeled dataset.  We then manually verified each PDF, keeping the 
pages that contain only solo instrumental sheet music and discarding any filler pages or pages containing accompaniment parts.  Some corrupted or mislabeled PDFs were also discarded.  In total, the labeled dataset contains 7627 valid pages of solo instrumental sheet music.  For the classification task, we split the labeled data by piece, using \(60\%\) for training, \(20\%\) for validation, and \(20\%\) for testing.  For language model training, we split the data by piece, using \(90\%\) for training and \(10\%\) for validation.

To construct data for the proxy task, we sampled 3600, 1200, and 1200 fragments from each instrument for train, validation, and test, respectively.  We construct separate datasets for fragments of size 64, 128, and 256.

\begin{figure}
	\centerline{\includegraphics[width=0.95\columnwidth]{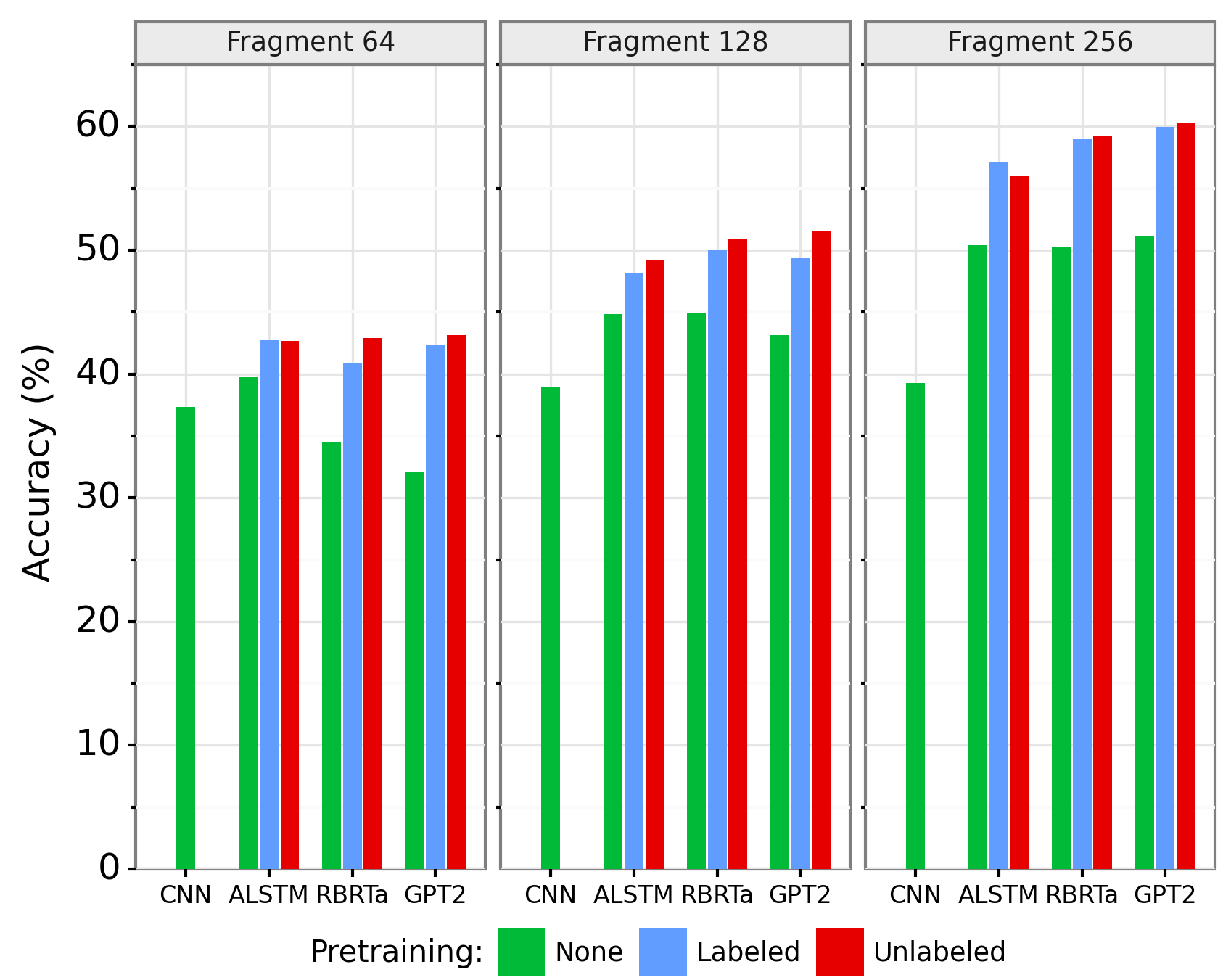}}
	\caption{ Results on the proxy classification task. This shows the effect of different pretraining conditions and fragment sizes.}
	\label{fig:resultsProxy}
\end{figure}

\section{Results}

We compare the performance of four different model architectures: CNN, AWD-LSTM, GPT-2, and RoBERTa.  Our CNN model is based on the model proposed in \cite{verma2019convolutional}, where local features are computed on a piano roll-like representation, the local features are pooled across time, and the resulting feature statistics are passed to a linear classifier.  Our model is not exactly the same as in \cite{verma2019convolutional}, however, since this previous work assumes that full symbolic music information is available.

Figure \ref{fig:resultsProxy} compares the performance of all models on the proxy classification task. We consider the performance for three different fragment sizes and three different pretraining conditions: no pretraining, labeled pretraining, and unlabeled pretraining.  With no pretraining, we train the classifier from scratch on the proxy data.  With labeled pretraining, we first train a language model on the labeled dataset, and then finetune the classifier on the proxy data.  With unlabeled pretraining, we train a language model on the unlabeled dataset, finetune the language model on the labeled data, and then finetune the classifier on the proxy data. 

\begin{figure}
	\centering
	\includegraphics[width=0.95\columnwidth]{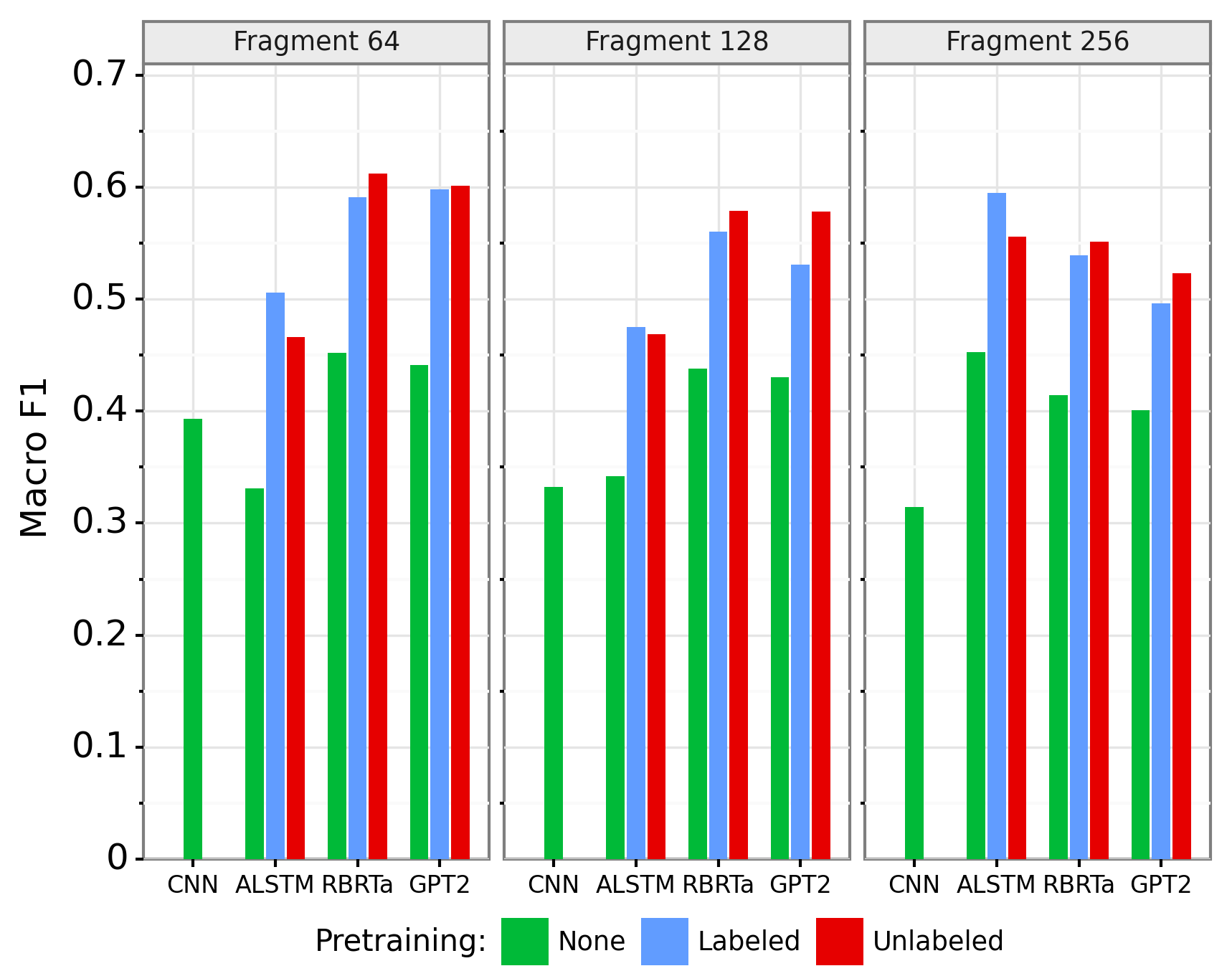}
	\caption{Model performance on fullpage classification task.}
	\label{fig:resultsFullPage}
\end{figure}

There are 3 things to notice about Figure \ref{fig:resultsProxy}.  First, language model pretraining helps a lot.  For all models and fragment sizes, we see a big increase in classification accuracy for the pretrained models (blue and red bars) compared to the models trained from scratch (green bars).  For example, the accuracy of the GPT-2 model with fragments of size 64 increases from \(32.1\%\) to \(42.3\%\) and \(43.2\%\) for the three pretraining conditions.  Second, there is only a small increase in classification accuracy going from labeled pretraining to unlabeled pretraining.  There are about 3 million bootleg score features in the labeled dataset and 5.6 million features in the unlabeled dataset.  Since this is only a modest increase in the amount of data and the unlabeled data is also noisy and imbalanced, the model does not benefit much from pretraining on the unlabeled data.  Finally, we see that the accuracy increases as the fragment size increases. This is to be expected, since the model has more context to work with.

Figure \ref{fig:resultsFullPage} shows the performance on the full page classification task. Note that the evaluation metric is macro F1 instead of accuracy, since the data in the full page classification task is imbalanced. We found averaging predictions on multiple fixed length sequences yielded superior results with all models except the CNN.  We only show the best-performing inference type for each model. 

There are two things to notice about Figure \ref{fig:resultsFullPage}.  First, we see that pretraining helps in all cases.  For all models and fragment sizes, the pretrained models (blue and red bars) perform much better than the models trained from scratch (green bars).  Second, for both transformer models the performance degrades as fragment size increases. 
For example, as the fragment size goes from 64 to 128 to 256, the macro F1 of the RoBERTa model decreases from 0.612 to 0.579 to 0.551.  We believe this is due to a bias in the proxy dataset: the proxy classifier is only trained on long fragments of size 256, but many pages of sheet music contain less than 256 words, resulting in data that the proxy classifier has not seen in training.  In contrast, the AWD-LSTM model seems to benefit from training on longer sequences, but is able to handle shorter sequences well even when short sequences are not seen in training.

\section{Analysis}
In this section, we perform three additional analyses on our best-performing model: RoBERTa with fragment size 64.

\begin{figure}
    \centerline{\includegraphics[width=.9\columnwidth]{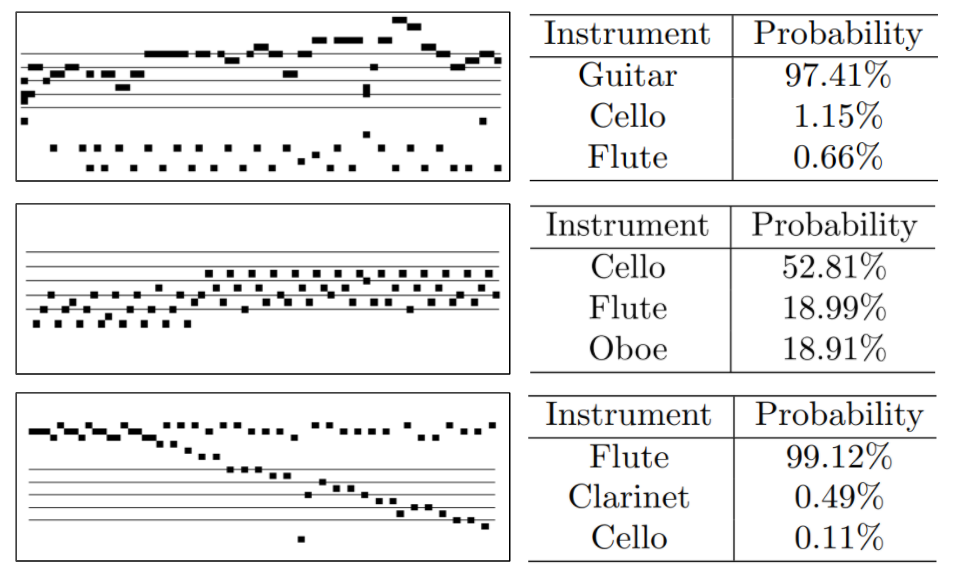}}
  
  \caption{Example outputs for the RoBERTa model.}
  \label{fig:sampleOutputs}
\end{figure}

The first analysis is to examine individual examples to better understand the model's behavior.  Figure \ref{fig:sampleOutputs} shows three bootleg score fragments (top - guitar, middle - guitar, bottom - flute) and the RoBERTa model predictions.  In the top example, the model confidently and correctly classifies the fragment as guitar, whereas it incorrectly classifies the middle example as cello.  In general, we found monophonic inputs to be much more error-prone than polyphonic inputs, since polyphony provides much more distinctive information about the instrument.  In the bottom example, the model confidently classifies the fragment as flute, recognizing that flute music frequently has high notes above the staff.

The second analysis is to examine common confusion pairs.  The most common (actual-predicted) confusion pairs are viola-flute (260), trumpet-oboe (259), and oboe-clarinet (230).  We noticed that the confusion pairs are often asymmetrical.  For example, there are far more oboe-clarinet confusions (230) than clarinet-oboe confusions (117).  We believe this asymmetry is due to the instrument ranges: because clarinet has a wider range than oboe, notes outside of the oboe's range provide information to avoid clarinet-oboe mistakes but not oboe-clarinet mistakes.  The instrument with the least mistakes is guitar, since it has the most polyphony.

\begin{figure}
    \centerline{\includegraphics[width=\columnwidth]{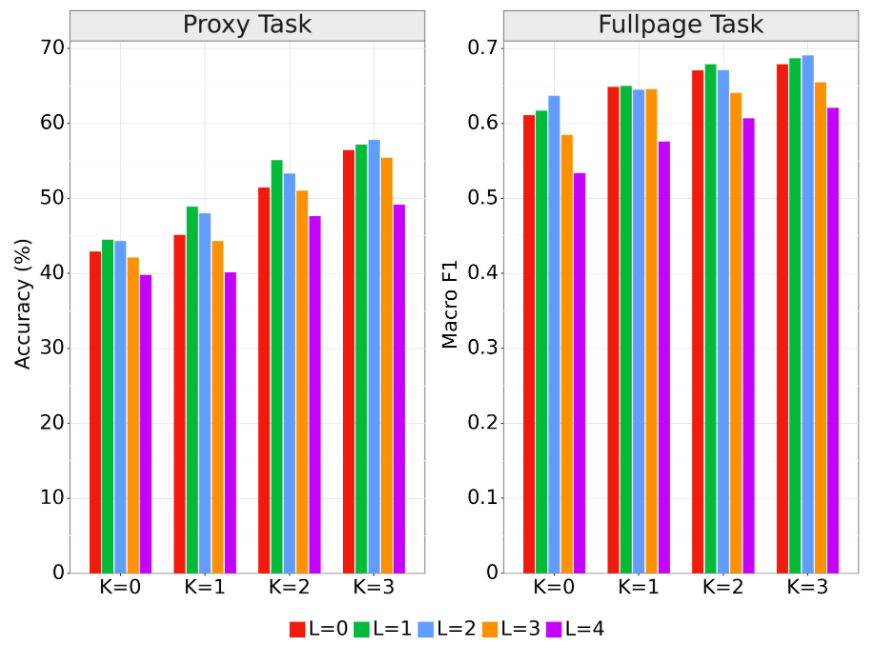}}
  \caption{Effect of data augmentation on the RoBERTa model.  Higher values of $K$ correspond to more training data augmentation, and higher values of $L$  correspond to more test-time augmentation.}
  \label{fig:effectAugmentation}
\end{figure}

The third analysis is to explore the effect of data augmentation.  We experimented with both training and test-time augmentation.  In training data augmentation, the bootleg scores are shifted up and down by up to $K$ staff line positions to generate additional data samples (e.g.~$K=3$ results in seven times more data).  Shifts correspond to a key transposition in the music.  In test time augmentation, inputs are shifted up and down by up to $L$ positions, and predictions are averaged across all shifted versions of the input.

Figure \ref{fig:effectAugmentation} shows the effect of data augmentation on the RoBERTa model.  We see that more training augmentation always helps up to $K=3$.  For example, the accuracy of the proxy classifier (with $L=0$) increases from 42.9\% to to 56.5\% as $K$ increases from 0 to 3, and the macro F1 score of the full page classifier increases from 0.611 to 0.679.  On the other hand, test-time augmentation only helps when used in moderation.  For example, the accuracy of the proxy classifier with $K=3$ improves from 56.5\% to 58.8\% as $L$ increases from 0 to 2.   Beyond $L=2$, however, we see a significant degradation in performance.  One possible explanation for this behavior is that large values of $L$ cause a loss of information about the range of the instrument.  Using the best configuration of $K=3$, $L=2$, the accuracy improves from 42.9\% to 58.8\% on the proxy task and from .611 to .691 macro F1 on the full page task.

\section{Conclusion}
We propose a method for predicting the instrument of a single page of solo instrumental sheet music. 
This method first converts the sheet music into a sequence of words and then feeds the words into a text classifier.  We demonstrate that language model pretraining substantially improves the performance of the classifier.  We also introduce two types of data augmentation that further improves classification performance.  For future work, we would like to incorporate clef detection into the system and expand our method to work with sheet music containing multiple instruments.

\section{Acknowledgments}
This work used the Extreme Science and Engineering Discovery Environment (XSEDE) Bridges at the Pittsburgh Supercomputing Center through allocation TG-IRI190019.  XSEDE is supported by National Science Foundation grant number ACI-1548562.  We also gratefully acknowledge the support of NVIDIA Corporation with the donation of the GPU used for training the models.

\vfill
\pagebreak

\bibliographystyle{IEEEbib}
\bibliography{InstrumentID}

\end{document}